\documentclass[runningheads]{llncs}

\PassOptionsToPackage{pagebackref}{hyperref}
\usepackage[final,year=2026,ID=8896]{eccv}

\usepackage{eccvabbrv}
\usepackage{graphicx}
\usepackage{booktabs}
\usepackage{amsmath}
\usepackage{amssymb}
\usepackage{multirow}
\usepackage[table,xcdraw]{xcolor}
\usepackage{algorithm}
\usepackage{algorithmic}
\usepackage[accsupp]{axessibility}
\usepackage{hyperref}
\usepackage{orcidlink}
\usepackage{stfloats}

\usepackage{color}
\newcommand{\xmli}[1]{}
\newcommand{\cz}[1]{}

\newcommand{\ours}{BBRD}

\newcommand{\ourslong}{Brightness Bias-Robust Denoising}

\newcommand{\numarch}{8}
\newcommand{\numbaseline}{13}

\begin{document}

\title{The Devil is in the Dark Pixels:\\ Toward Brightness Bias-Robust Denoising}
\titlerunning{Brightness Bias in Learned Image Denoising}
\author{Sungjun Cho\inst{1} \and
Zhuangzhuang Chen\inst{1} \and
Xiaomeng Li\inst{1}\thanks{Corresponding author.}}
\authorrunning{S. Cho et al.}
\institute{Department of Electronic and Computer Engineering,\\
The Hong Kong University of Science and Technology, Hong Kong SAR, China\\
\email{schoaq@connect.ust.hk, eezzchen@ust.hk, eexmli@ust.hk}}
\maketitle

\begin{abstract}
In this paper, we reveal an important yet overlooked problem in image
denoising: under signal-dependent camera noise models, dark regions
suffer from inherently low Signal-to-Noise Ratio (SNR), as signal
intensity decays far faster than noise variance diminishes, making
detail recovery in dark areas fundamentally challenging.
Yet rather than compensating for this difficulty, MSE-trained
denoisers \emph{exacerbate} it---reconstructing dark pixels up to
$6{\times}$ worse relative to their per-band noise floor.
This bias stems from two compounding factors: signal-dependent noise
inflates bright-pixel residuals, and the network's Jacobian norm
increases monotonically with brightness. Together, these cause bright
regions to chronically dominate gradient updates at the expense of
dark ones. To this end, we propose \emph{\ourslong{}} (\ours{}), a drop-in
replacement for MSE loss that partitions pixels into brightness bands,
normalizes per-band error by empirical noise variance, and applies
Group Distributionally Robust Optimization (Group-DRO) to dynamically
upweight whichever band is currently worst,
with zero additional parameters or inference cost.
Across \numarch{} architectures and 2 datasets in our experiments,
\ours{} is the \textbf{only method among \numbaseline{} tested
alternatives that improves each brightness band simultaneously},
achieving up to +0.45 dB on dark bands, +0.32 dB on bright bands, and +0.65 dB aggregate Peak Signal-to-Noise Ratio (PSNR) on SIDD, with the largest per-band gains in the darkest regions where detail recovery matters most.
Code is available at \url{https://github.com/xmed-lab/BBRD}.



\keywords{Image denoising \and Brightness bias \and
Signal-dependent noise \and Distributionally robust optimization}
\end{abstract}

\section{Introduction}
\label{sec:intro}

\begin{figure}[t]
  \centering
  \begin{subfigure}{0.9\linewidth}
    \centering
    \includegraphics[width=\linewidth]{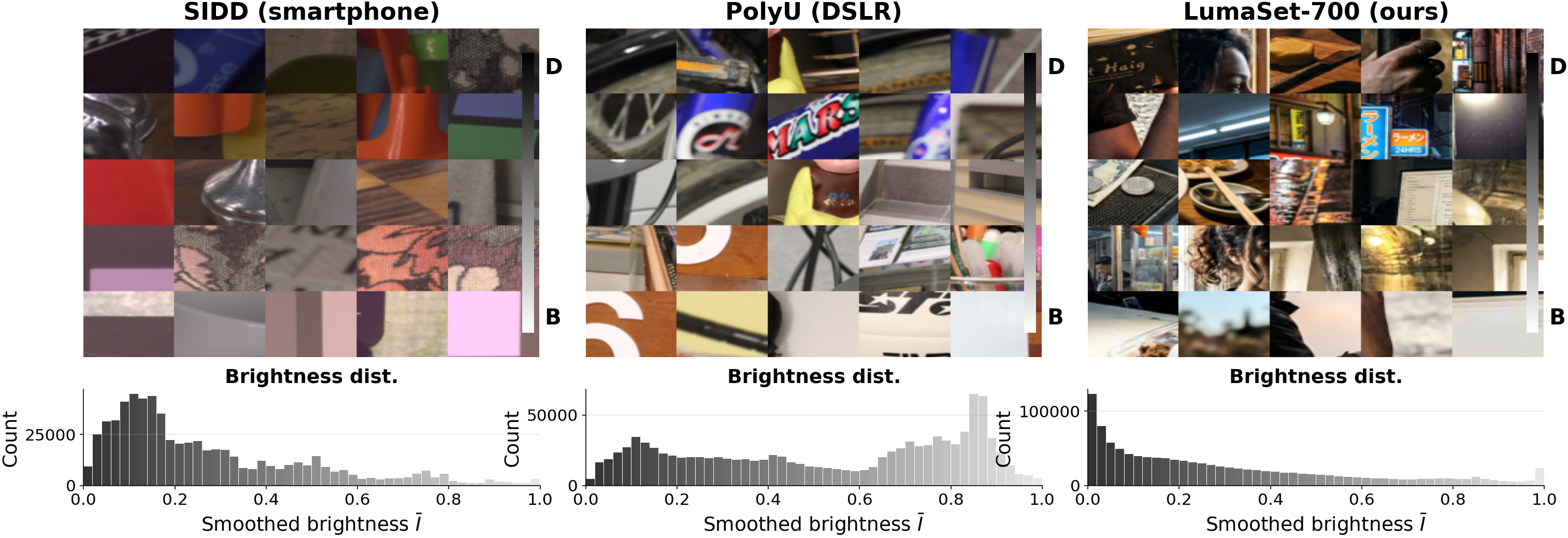}
    \caption{Brightness distribution of SIDD, PolyU, and our dataset.}
    \label{fig:brightness_dist}
  \end{subfigure}
  \vspace{0.3em}
  \begin{subfigure}{0.8\linewidth}
    \centering
    \includegraphics[width=\linewidth]{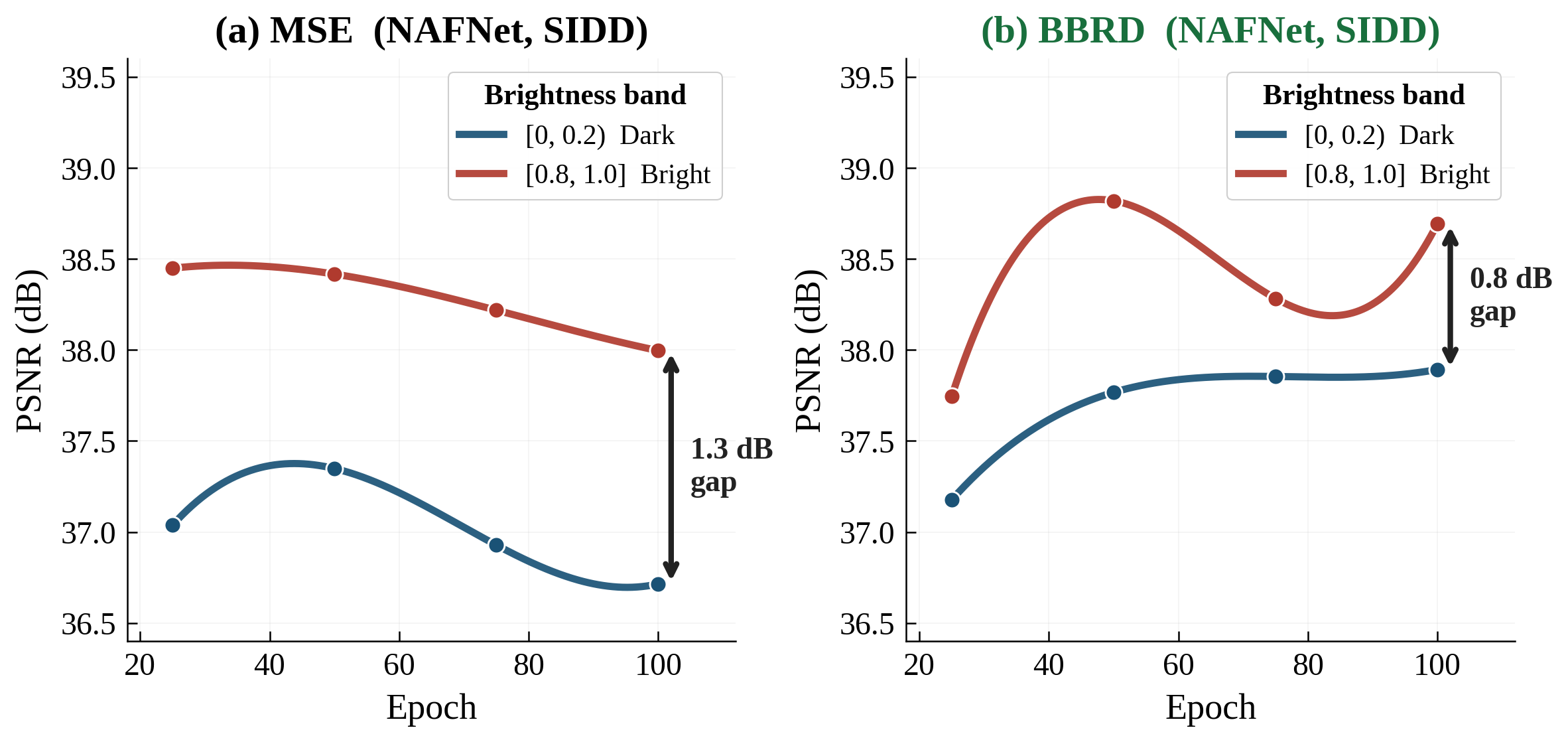}
    \caption{Per-band PSNR during training (NAFNet, SIDD).
    Both panels use the same model, dataset, and brightness bands;
    only the loss differs (MSE vs.\ \ours{}).}
    \label{fig:overview}
  \end{subfigure}
  \caption{%
    \textbf{(a)} Brightness distribution across datasets.
    \textbf{(b)} Under identical NAFNet training on SIDD dataset,
    MSE under-optimizes dark regions (1.3\,dB gap at convergence);
    \ours{} reduces this to 0.8\,dB with zero additional parameters.}
  \label{fig:tradeoff}
\end{figure}

\begin{figure*}[t]
  \centering
  \begin{subfigure}{0.35\linewidth}
    \centering
    \includegraphics[width=\linewidth]{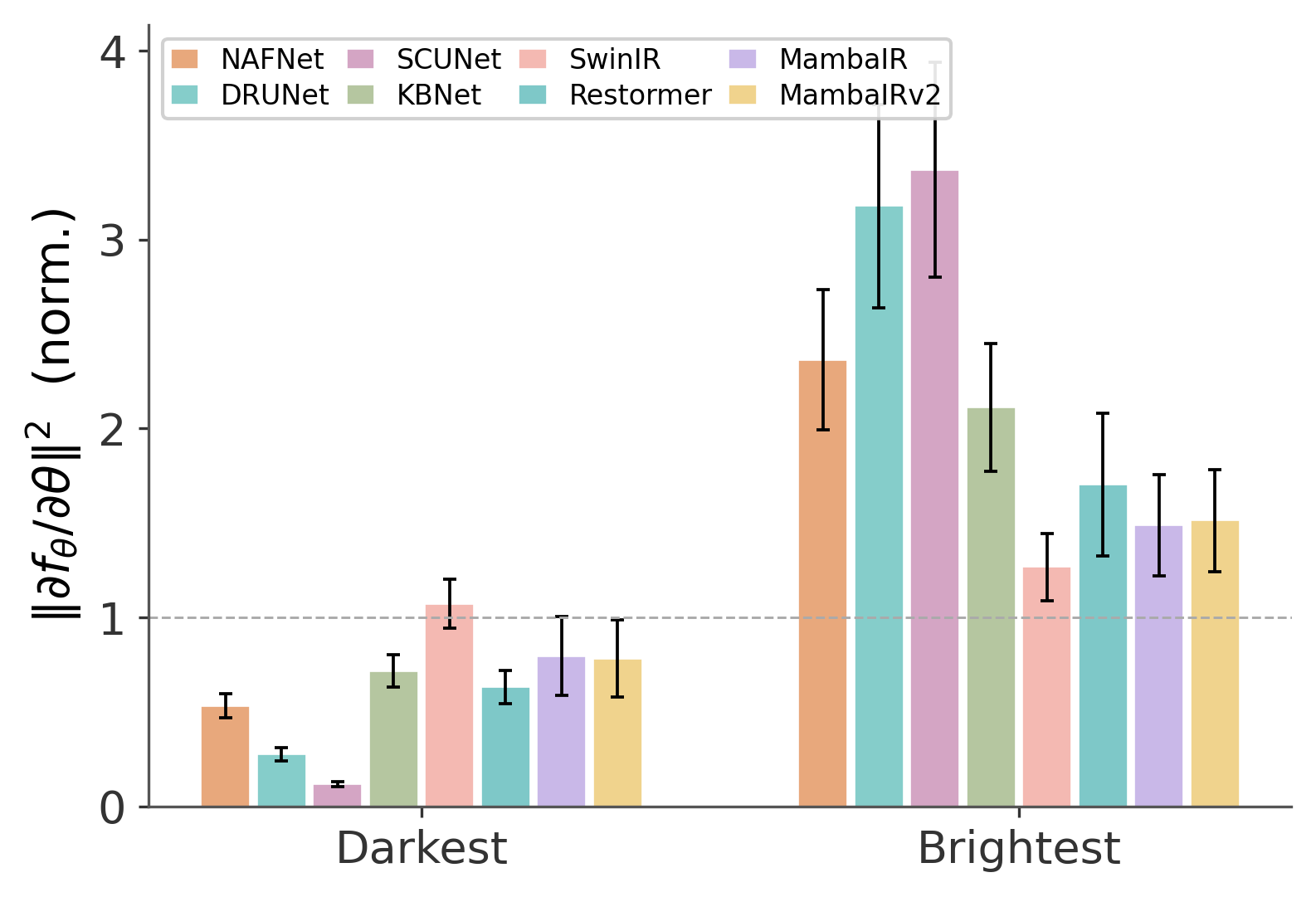}
    \caption{Jacobian norm on SIDD}
    \label{fig:overview1}
  \end{subfigure}
  \hfill
  \begin{subfigure}{0.35\linewidth}
    \centering
    \includegraphics[width=\linewidth]{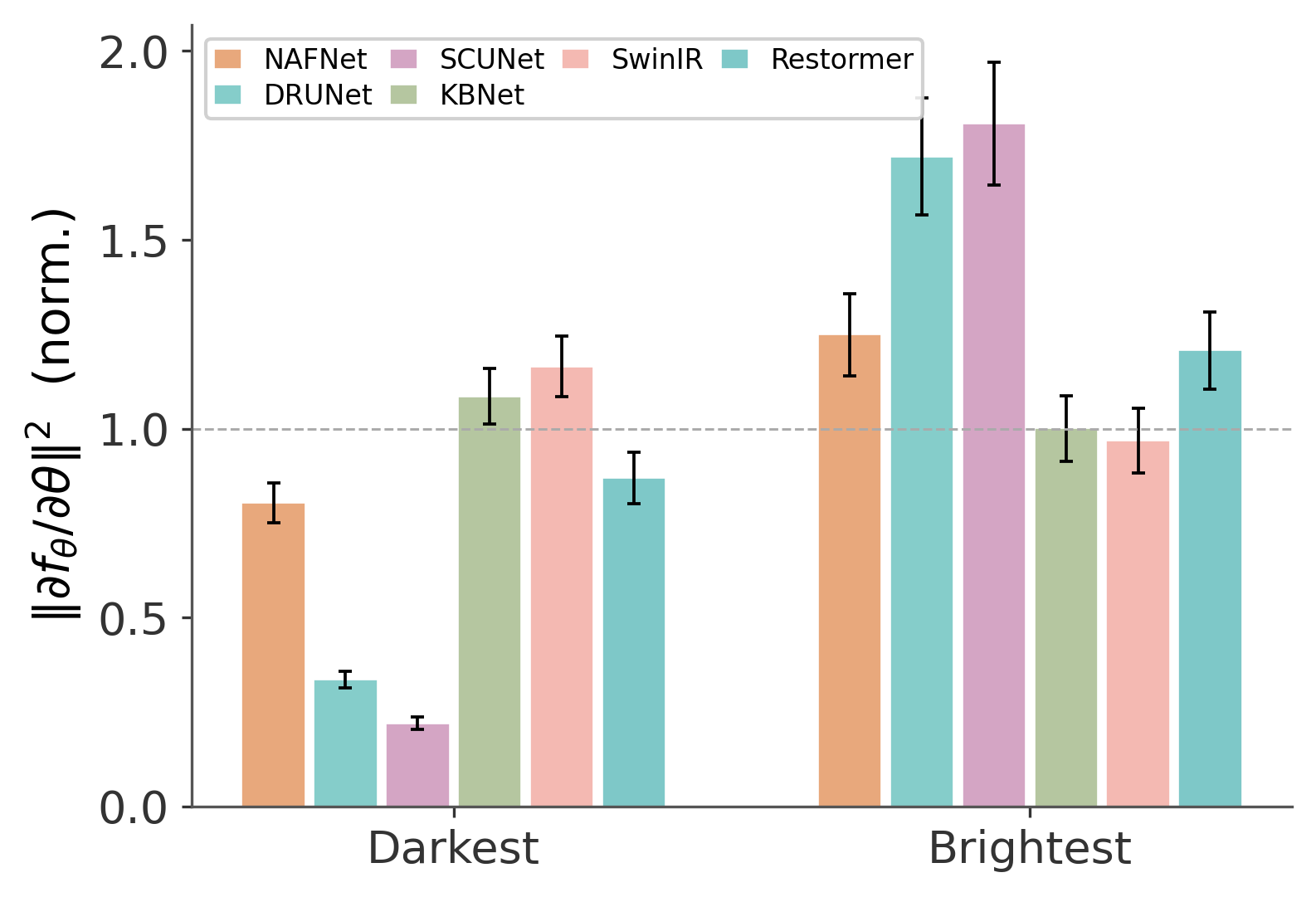}
    \caption{Jacobian norm on PolyU}
    \label{fig:overview2}
  \end{subfigure}
  \hfill
  \begin{subfigure}{0.25\linewidth}
    \centering
    \includegraphics[width=\linewidth]{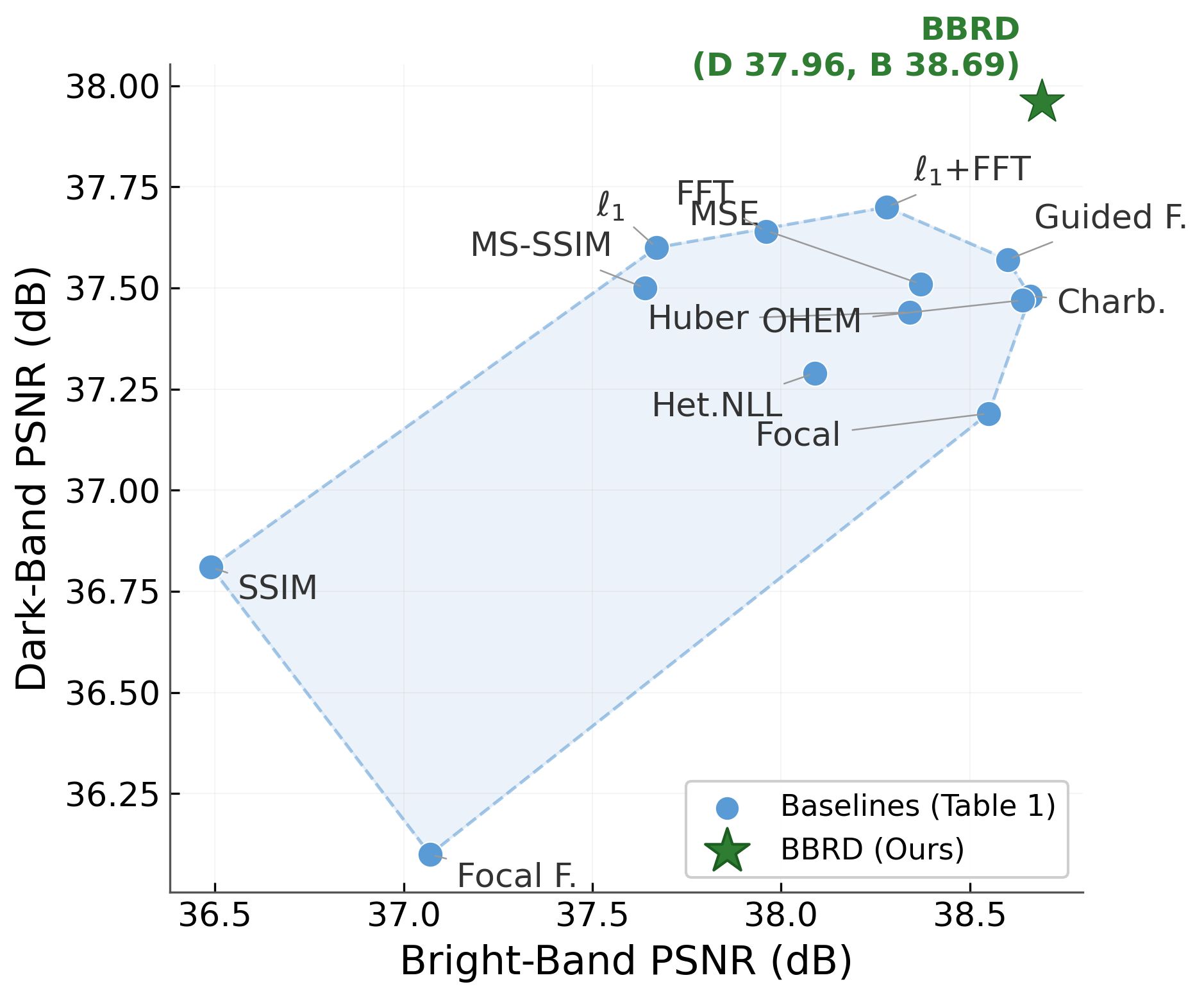}
    \caption{SIDD}
    \label{fig:pareto_sidd}
  \end{subfigure}
  \caption{%
    Jacobian norm of the brightest band consistently exceeds that
    of the darkest band under MSE across
    \numarch{} architectures (CNNs, Transformers, SSMs) on
    \textbf{(a)}~SIDD and \textbf{(b)}~PolyU,
    confirming the gradient imbalance is architecture-agnostic.
    \textbf{(c)}~\textcolor{green}{\ours{}} (PSNR-D\,37.96, PSNR-B\,38.69)
    strictly dominates all baselines---including the strongest,
    $\ell_1$+FFT (37.70, 38.28) and MSE (37.51, 38.37)---pushing the
    Pareto frontier outward on SIDD with zero additional parameters.}
  \label{fig:jacobian_frontier}
\end{figure*}

Deep learning has driven rapid progress in image denoising~\cite{elad2023survey, jiang2025survey}, with architectures advancing from CNNs~\cite{dncnn,drunet,nafnet} through Transformers~\cite{swinir,restormer} to State-Space Models~\cite{mambair,mambairv2}, finding broad use in photography, medical imaging, and autonomous driving~\cite{Chen_2023_ICCV,sun2025ntire, li2025aim, pathak2026medical}.
Despite this architectural diversity, all models share a common training paradigm: per-pixel MSE or $\ell_1$ minimization, evaluated by a single aggregate PSNR.
We reveal an important yet overlooked failure of this paradigm: \emph{brightness bias}.
Under a Poisson--Gaussian noise model, although absolute noise variance in dark regions is small, signal intensity decays far more rapidly, yielding inherently low Signal-to-Noise Ratio (SNR) in dark areas~\cite{pmcoverview2025}.
As shown in \cref{fig:overview}, under MSE training the dark band decays to 36.7\,dB while the bright band remains near 38.0\,dB---a \textbf{1.3\,dB gap that persists and
even widens} as training progresses.
At convergence, the darkest band of NAFNet~\cite{nafnet} on SIDD
achieves only 37.51\,dB versus 38.37\,dB for the brightest band,
and exhibits $6{\times}$ higher normalized error $R_k$ relative
to its per-band noise floor. This excess gap 
cannot be explained by SNR alone, pointing to a systematic 
failure in how these models are trained.
This raises a fundamental question:
\textit{\textbf{Why do learned denoisers disproportionately 
under-optimize dark regions, compounding their already 
unfavorable SNR conditions?}}
To answer this question, we conduct a series of experiments that train various denoising 
models on the SIDD dataset. As shown in \cref{fig:overview1,fig:overview2}, 
the Jacobian norm of the brightest band (B5) consistently exceeds that of the darkest band (B1)
across all \numarch{} architectures spanning
CNNs (NAFNet, DRUNet, SCUNet, KBNet),
Transformers (SwinIR, Restormer), and
State Space Models (MambaIR, MambaIRv2).
This confirms that \emph{brightness bias} is not 
attributable to model capacity or architectural inductive bias
but to the intrinsic properties of the MSE loss function. 
The reason is twofold: (1) Under signal-dependent noise, brighter pixels naturally yield larger residuals, 
and (2) brighter pixels induce larger Jacobian norms. Intuitively, higher input activations produce larger intermediate feature magnitudes, which---after passing through any monotone or near-monotone nonlinearity (ReLU, GELU, SiLU, SimpleGate)---yield larger partial derivatives and thus amplify gradient flow. This is empirically confirmed across all \numarch{} architectures despite their diverse activation choices (\cref{fig:overview1,fig:overview2}). These two factors compound to dominate gradient updates, meaning the network's weights are updated primarily to fit bright regions. 
This gradient imbalance leaves dark regions chronically under-optimized (\cref{sec:theory}). 
This misalignment between what the loss optimizes and what the image actually needs reveals a 
fundamental limitation of uniform pixel-wise objectives in the presence of signal-dependent noise.

In light of this analysis, a straightforward solution is to adopt existing reweighting loss functions 
to focus on dark pixels. However, difficulty-based reweighting methods such as Focal Loss~\cite{focalloss} and Online Hard Example Mining (OHEM)~\cite{ohem} mistakenly identify bright pixels as ``hard'' simply because they are signal-dependent, thus yielding larger residuals. As a result, these methods inadvertently \emph{amplify} the bias by unintentionally doubling down on the existing gradient imbalance. Other objectives, e.g., robust pixel losses, uncertainty modeling, and frequency-domain losses, either operate uniformly over all pixels or fail to track which brightness band is currently lagging (comparison results can be found in \cref{sec:main_results}). Moreover, there is no standard protocol to measure brightness disparity, and aggregate PSNR obscures the gap entirely, making the problem invisible to conventional evaluation.

To address these issues, we make the following contributions.
First, we introduce a standardized brightness-disparity evaluation protocol
for image denoising: per-band PSNR over five brightness strata,
validated across \numarch{} architectures and \numbaseline{} loss
functions.
Second, we propose \emph{\ourslong{}} (\ours{}), a drop-in MSE
replacement that corrects brightness bias via data-driven band
partitioning, per-band noise normalization, and softmax Group DRO, with
zero additional parameters or inference cost. Unlike existing losses that trade off between dark and bright performance, \ours{} pushes the Pareto frontier outward.
As shown in \cref{fig:jacobian_frontier}, in our experiments \ours{}
is the \textbf{only objective among \numbaseline{} tested alternatives
that improves every brightness band simultaneously}, achieving
up to \textbf{+0.65\,dB} aggregate PSNR on SIDD with visibly sharper
dark-region textures (\cref{fig:qualitative}). 
Finally, as existing benchmarks lack balanced brightness coverage, we introduce \textbf{LumaSet-700}, a synthetic evaluation benchmark with controlled noise, and show that \ours{} generalizes beyond signal-dependent noise.
Our main contributions are summarized as follows:
\begin{itemize}
     \item We introduce \ours{}, a drop-in MSE replacement that corrects
    brightness bias via data-driven band partitioning, per-band noise
    normalization, and softmax Group DRO, achieving up to +0.65\,dB on SIDD
    with zero additional parameters or inference cost (\cref{sec:method}).
    \item We propose the first standardized per-band brightness-disparity
    evaluation protocol, demonstrating that brightness bias is universal
    across \numarch{} architectures, 2  real-world datasets, and \numbaseline{} loss
    functions (\cref{sec:exp}).
    \item We introduce \textbf{LumaSet-700}, a synthetic 
    benchmark of 700 test patches stratified across dark, mid-tone, and
    bright scenes with controllable noise, demonstrating that \ours{}
    generalizes beyond signal-dependent noise (\cref{sec:noise_robustness}).
    
\end{itemize}
\section{Related Work}
\label{sec:related}

\noindent\textbf{Deep image denoising.}
Deep image denoising has progressed through four major paradigms, each
advancing architectural capacity while sharing a common training convention.
CNN-based methods such as DnCNN~\cite{dncnn}, DRUNet~\cite{drunet},
and NAFNet~\cite{nafnet} established strong baselines using MSE or $\ell_1$
loss. Transformer-based models including SwinIR~\cite{swinir} and
Restormer~\cite{restormer} extended the receptive field via attention
mechanisms, adopting Charbonnier loss for denoising.
Diffusion models~\cite{ho2020ddpm,diffbir,wu2024osediff,dong2025tsdsr}
offer strong perceptual quality through iterative or distilled score-based
refinement, but are trained by predicting added noise at each timestep,
an objective equivalent to per-noise-level MSE with uniform pixel weighting.
Most recently, State Space Models such as MambaIR~\cite{mambair} and
MambaIRv2~\cite{mambairv2} achieve linear-complexity global modeling, yet continue training with Charbonnier loss for denoising.

Notably, the above methods share the same critical blind spot: their base reconstruction objectives treat every pixel uniformly regardless of brightness, and are evaluated solely by aggregate PSNR, which renders brightness-stratified disparities entirely invisible. More importantly, the dominant trend in image denoising research
has been to improve model architecture---from CNNs to Transformers
to State Space Models---while keeping the MSE loss fixed.
\ours{} takes the orthogonal direction: it keeps the model
unchanged and corrects the loss, making it directly compatible
with and validated on the latest SOTA architectures including
MambaIR and MambaIRv2.

\noindent\textbf{Loss functions for bias-robust learning.}
MSE ($\ell_2$) remains the default for PSNR-oriented training.
Alternative pixel-wise objectives like $\ell_1$ and Charbonnier
loss~\cite{charbonnier1994} are adopted in SwinIR~\cite{swinir} and
Restormer~\cite{restormer} for sharper gradient behavior, while Huber
loss~\cite{huber1964} improves outlier robustness but does not account
for structured noise heteroscedasticity.
To address varying difficulty, Focal Loss~\cite{focalloss} and
OHEM~\cite{ohem} assign higher weights to harder examples; however,
under signal-dependent noise, ``hard'' examples concentrate in
high-variance bright regions, so these methods inadvertently
\emph{amplify} brightness bias rather than correcting it.
Heteroscedastic NLL~\cite{kendall2017} models per-pixel aleatoric
uncertainty by jointly learning $\sigma(x)$, yet merely confirms the
bias by assigning low uncertainty to dark bands whose noise variance is
genuinely low.
Finally, SSIM~\cite{ssim2004}, perceptual losses~\cite{perceptualloss},
and frequency-domain losses~\cite{wavelettransform,fftloss,msssim}
operate uniformly over all pixels and are entirely blind to
brightness-dependent noise structure.

Herein, the key difference of \ours{} is that we explicitly
normalize per-band error by empirical noise variance before reweighting,
which is the critical step that transforms reweighting from a bias-amplifier into a bias-corrector.

\noindent\textbf{Distributionally robust optimization.}
Group DRO~\cite{groupdro} minimizes worst-group loss in classification
and has strong convergence guarantees.
However, applying Group DRO directly to image restoration requires
non-trivial adaptation: unlike classification where groups are defined
across samples, a single image patch spans multiple brightness levels,
requiring groups to be defined at the pixel level within each patch.
Furthermore, without noise normalization, naive Group DRO
\emph{worsens} brightness bias, as it mis-identifies bright bands as
worst-performing because their raw MSE is inflated by signal-dependent
noise, and upweights them further, compounding the exact problem we
aim to fix.

In contrast, \ours{} first normalizes per-band errors by empirical
noise variance, ensuring the worst-performing band is identified on a
noise-corrected scale rather than by raw loss magnitude, and then
replaces the non-differentiable minimax objective with softmax DRO to
enable dynamic reweighting during training.
Without normalization, DRO upweights bright bands (highest raw loss);
with it, DRO correctly upweights dark bands (highest noise-normalized loss).


\section{Method}
\label{sec:method}

\subsection{Preliminaries}
\vspace{-2pt}
\noindent\textbf{Group Distributionally Robust Optimization.}
Standard ERM minimizes the average loss across all training samples,
which can leave minority groups severely under-optimized.
Group DRO~\cite{groupdro} addresses this by minimizing the
worst-group loss instead:
\begin{equation*}
    \min_\theta \max_{g \in \mathcal{G}} \mathcal{L}_g(\theta),
\end{equation*}
where $\mathcal{G}$ is a predefined set of groups and
$\mathcal{L}_g$ is the loss on group $g$.
This ensures no group is left behind at the expense of aggregate
performance.

\subsection{Why MSE Fails: Two Compounding Factors}
\label{sec:theory}
Although MSE is widely adopted, its uniform pixel treatment leads to a systematic gradient imbalance under signal-dependent noise. Two independent factors \emph{compound} to over-emphasize bright pixels during training:
\begin{itemize}
    \item \textbf{Signal-dependent residuals.} In early-to-mid training the residual approximates the noise, $r_i \approx n_i$, so $\mathbb{E}[r_i^2] \approx \mathrm{Var}(n_i)$ grows with brightness. Brighter pixels therefore produce larger gradient contributions via Eq.~\eqref{eq:gradient} by default, before any network-specific effects.
    \item \textbf{Brightness-correlated Jacobian.} The network Jacobian $\|\partial f_\theta(x_i)/\partial\theta\|$ is empirically larger for high-activation (bright) inputs across all \numarch{} evaluated architectures---CNNs, Transformers, and State Space Models (\cref{fig:overview1,fig:overview2})---independently amplifying bright-pixel gradients.
\end{itemize}

\noindent\textbf{Gradient analysis.}
The MSE gradient at pixel $i$ is:
\begin{equation}
    \frac{\partial \mathcal{L}}{\partial \theta}\bigg|_i =
    \frac{2}{N}(f_\theta(x_i) - y_i) \cdot
    \frac{\partial f_\theta(x_i)}{\partial \theta}.
    \label{eq:gradient}
\end{equation}

\noindent\textbf{Proposition 1} (Gradient imbalance).
\textit{Under the following assumptions:
\textbf{(A1)} residuals approximate noise in early-to-mid training
($r_i \approx n_i$);
\textbf{(A2)} noise variance $\mathrm{Var}(n_i)$ and Jacobian norm
$\|\partial f_\theta(x_i)/\partial\theta\|$ both increase
with brightness (empirically confirmed across all
\numarch{} architectures in \cref{fig:overview1,fig:overview2}),
the expected squared gradient magnitude increases monotonically
with brightness:}
\begin{equation}
    \mathbb{E}\!\left[\left|\frac{\partial \mathcal{L}}{\partial
    \theta}\bigg|_i\right|^2\right]
    \;\propto\;
    \underbrace{\mathbb{E}[r_i^2]}_{\text{factor (i)}}
    \cdot
    \underbrace{\mathbb{E}\!\left[\left\|\frac{\partial f_\theta(x_i)}
    {\partial \theta}\right\|^2\right]}_{\text{factor (ii)}},
    \label{eq:prop1}
\end{equation}
\textit{causing bright pixels to dominate parameter updates and
leaving dark regions chronically under-optimized.}

\noindent\textit{Proof.}
From Eq.~\eqref{eq:gradient}, the squared gradient at pixel $i$
is proportional to $r_i^2 \cdot \|J_i\|^2$.
By (A1), $\mathbb{E}[r_i^2] \approx \mathrm{Var}(n_i)$
increases with brightness.
By (A2), $\mathbb{E}[\|J_i\|^2]$ also increases with brightness.
Since both non-negative factors increase monotonically with
brightness, their product likewise increases monotonically,
completing the argument.\hfill$\square$

\noindent\textbf{Jacobian measurement.}
$\|\partial f_\theta(x_i)/\partial\theta\|^2$ is approximated as
$\sum_{p} \|\nabla_{\theta}^{(p)}\|^2_F$,
the sum of squared Frobenius norms over all trainable parameter
tensors $p$, computed via a single \texttt{backward()} pass on
the per-pixel (or per-band) output scalar.
Per-band values are obtained by zeroing all output pixels outside
band $k$ before calling \texttt{backward()}, so that autograd
accumulates gradients only from band-$k$ pixels; the resulting
$\sum_p\|\nabla_\theta^{(p)}\|^2_F$ is then divided by
$|\mathcal{B}_k|$.
No additional normalization is applied; \cref{fig:overview1,fig:overview2}
report these \emph{unnormalized} per-band Jacobian norms as the
primary result.
As a sanity check, we also recompute norms normalized by parameter
count per layer,
$\tilde{J}_i = \sum_p \|\nabla_{\theta}^{(p)}\|^2_F / |\theta^{(p)}|$,
and confirm the same trend across all
\numarch{} architectures---including Transformer-based models
(SwinIR, Restormer), whose self-attention layers could in
principle redistribute activation energy, and State Space Models
(MambaIR, MambaIRv2): the absolute scale changes but the
brightest band consistently dominates the darkest in every case,
confirming that the phenomenon is tied to brightness, not to
layer size or architectural inductive bias.
\noindent\textbf{Persistent bias.}
This imbalance is self-reinforcing: once the network enters the
fine-convergence regime ($r_i \ll n_i$), dark-pixel gradients are
too weak to undo the accumulated deficit (\cref{fig:dynamics_mse}),
and any reweighting of raw MSE inherits this bias since the gradient
signal is already corrupted by noise heteroscedasticity.

\subsection{Brightness Bias-Robust Denoising (\ours{})}

\begin{figure*}[t]
  \centering
  \includegraphics[width=0.99\textwidth]{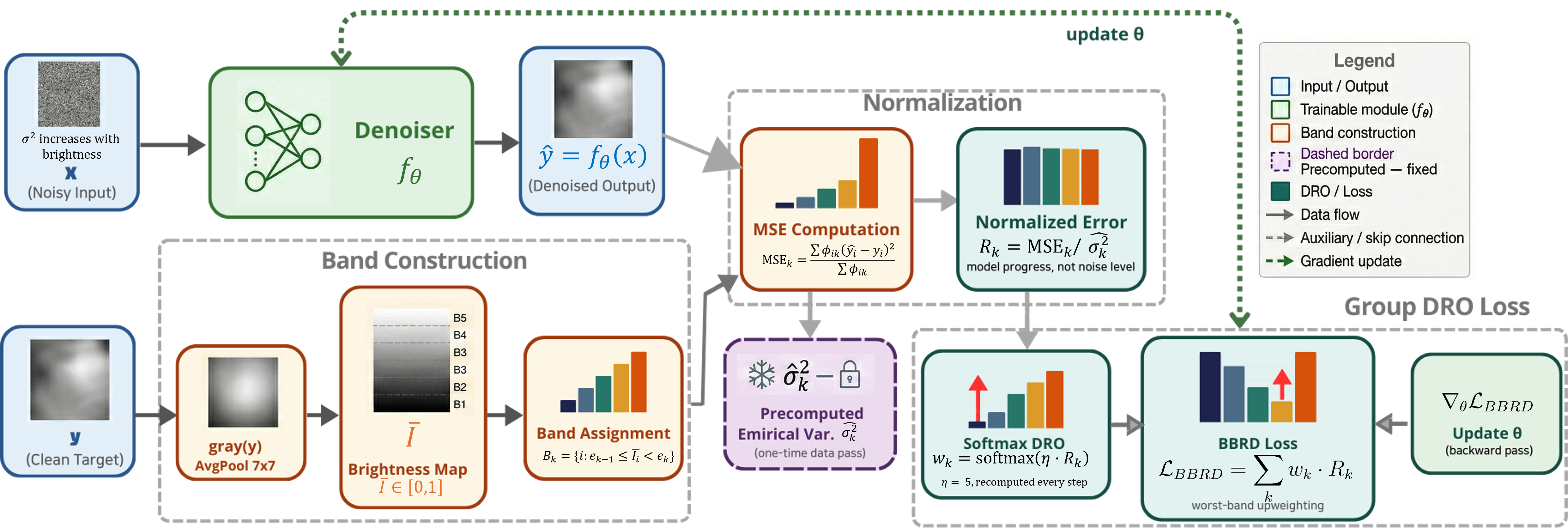}
  \caption{\textbf{\ours{} training pipeline.}
  A smoothed brightness map partitions pixels into $K$ bands via GMM and
  Gaussian soft assignments (orange). Per-band MSE is normalized by empirical
  noise variance $\hat{\sigma}_k^2$, yielding $R_k$ (teal). Softmax DRO
  ($\eta{=}5$) upweights the worst band, forming
  $\mathcal{L}_\text{BBRD}=\sum_k w_k R_k$ (dark teal).
  Band computation is training-only; inference runs $f_\theta$ with zero overhead.}
  \label{fig:pipeline}
\end{figure*}

\noindent\textbf{Overview:}
As illustrated in \cref{fig:pipeline}, \ours{} corrects brightness
bias via three sequential steps, each necessary: removing any one
breaks the correction (\cref{tab:ablation}).
In the first step, we partition pixels into brightness-specific bands
via a data-driven GMM with Gaussian soft assignments, enabling the
loss to track brightness-specific failures smoothly across the full
intensity range.
In the second step, we normalize per-band error by empirical noise
variance, placing all bands on a noise-corrected scale so that the
truly worst-performing band, not merely the noisiest, is correctly
identified.
In the third step, we apply Softmax Group DRO to dynamically upweight
whichever band is currently lagging, continuously redirecting
optimization pressure throughout training. The result is a zero-parameter, drop-in replacement for MSE with no inference overhead (\cref{fig:pipeline}).

\noindent\textbf{Brightness Band Construction.}
Brightness bias manifests differently across the intensity spectrum:
dark bands suffer from both low gradient magnitude and low noise
variance, while bright bands dominate training due to the opposite.
A single uniform MSE cannot track these band-specific failure modes;
to redirect optimization pressure where it is truly needed, we must
first partition pixels into brightness-specific groups so that
per-band progress can be measured and compared on a common scale.

\textbf{Brightness map.} For a clean patch $y$, we compute a smoothed brightness map $\bar{I} = \mathrm{AvgPool}_{7\times7}(\mathrm{gray}(y))$ $\in [0,1]^{H\times W}$ using sRGB weights $(0.299, 0.587, 0.114)$.
Bands are defined on the clean target $y$ rather than the noisy
input, avoiding noise-induced misassignments; this map is
computed only during training, incurring no inference overhead.

\textbf{GMM fitting and BIC selection.}
We fit a 1-D GMM to training-set intensities and select $K^*$ via
BIC, which balances fit quality against model complexity by penalizing
the number of components.
Unlike uniform binning, which arbitrarily splits the intensity range
into equal intervals, GMM identifies where pixels naturally cluster
in brightness space, placing band boundaries at the valleys between
adjacent components where pixel density is lowest.
This data-adaptive partitioning ensures each band captures a
coherent group of pixels with similar brightness characteristics,
rather than mixing pixels from different luminance modes.
BIC then automatically selects the number of bands that best
describes the data without overfitting, selecting $K^*{=}5$ for
SIDD and $K^*{=}3$ for PolyU without manual tuning
(\cref{sec:component_ablation}). Then, we propose \textbf{Gaussian soft assignment.}
Hard binning introduces gradient discontinuities at boundaries,
where a small change in $\bar{I}_i$ abruptly switches band
membership and can cause training instability.
We instead assign pixel $i$ to band $k$ via a Gaussian soft weight:
\begin{equation}
    \phi_{ik} = \exp\!\left(-\frac{(\bar{I}_i - c_k)^2}{2\sigma_g^2}\right),
    \quad c_k = \frac{e_{k-1}+e_k}{2},
    \label{eq:gaussian_weight}
\end{equation}
where $e_{k-1}$ and $e_k$ denote the valley boundaries flanking
band $k$ (i.e., the intensity minima between adjacent GMM components),
$c_k$ is the band center, and $\sigma_g{=}0.05$ ensures
overlapping support near boundaries, so pixels near a boundary
contribute smoothly to both adjacent bands.
Note that $\phi_{ik}$ are unnormalized weights; normalization
is implicit via the denominator $\sum_i \phi_{ik}$ in
Eq.~\eqref{eq:soft_mse}. Per-band MSE is then:
\begin{equation}
    \mathrm{MSE}_k = \frac{\sum_i \phi_{ik}\,(\hat{y}_i - y_i)^2}{\sum_i \phi_{ik}}.
    \label{eq:soft_mse}
\end{equation}

\begin{figure*}[ht]
  \centering
  \includegraphics[width=0.99\linewidth]{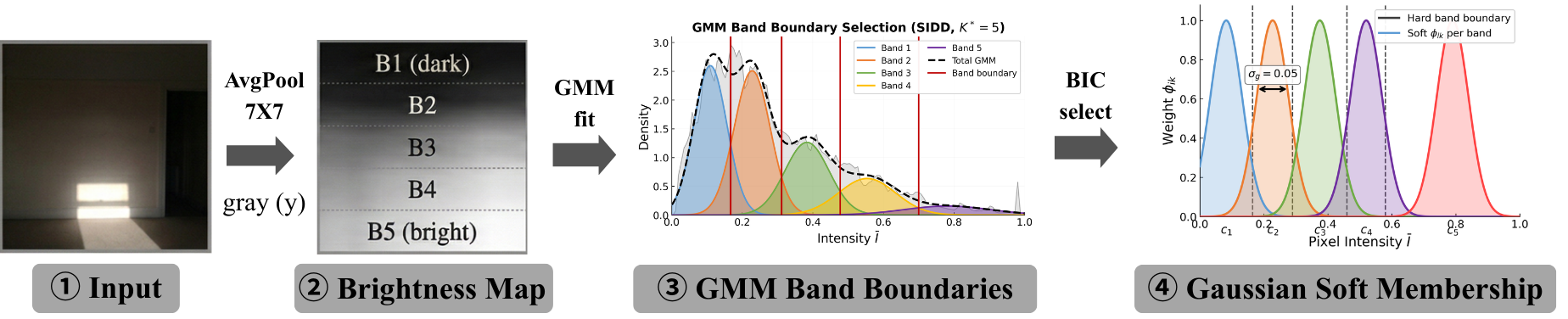}
  \caption{\textbf{Brightness band construction.}
  \textbf{(1)}~Input patch.
  \textbf{(2)}~Smoothed brightness map $\bar{I}$ via AvgPool.
  \textbf{(3)}~GMM fitted to training-set intensities; boundaries placed at
  inter-component valleys ($K^*{=}5$ for SIDD, $K^*{=}3$ for PolyU).
  \textbf{(4)}~Gaussian soft weights $\phi_{ik}$ ensure smooth pixel-to-band assignment.}
  \label{fig:band_construction}
\end{figure*}

\noindent\textbf{Empirical Noise Normalization.}
Without normalization, brighter bands always exhibit higher raw MSE
simply due to signal-dependent noise, causing any reweighting scheme
to misidentify them as the worst-performing and further amplify the
bias we aim to correct.
We normalize with per-band empirical noise variance $\hat{\sigma}_k^2$,
precomputed in a single pass over the training set \emph{before}
training begins:
\begin{equation}
    \hat{\sigma}_k^2 = \frac{\sum_i \phi_{ik}\,(x_i - y_i)^2}{\sum_i \phi_{ik}},
    \qquad
    R_k = \frac{\mathrm{MSE}_k}{\hat{\sigma}_k^2}.
    \label{eq:norm}
\end{equation}
\textbf{Interpretation:} The normalized ratio $R_k$ has a natural
interpretation: $R_k \approx 1$ means no denoising progress;
$R_k \to 0$ is perfect reconstruction. This requires no parametric
noise model and adapts to any noise distribution. Since
$\hat{\sigma}_k^2$ is a data property rather than a model parameter,
it remains fixed throughout training.

\textbf{Numerical stability:} For very dark bands where
$\hat{\sigma}_k^2$ is small, dividing by $\hat{\sigma}_k^2$ could
amplify noise in $R_k$. We therefore clip $\hat{\sigma}_k^2$ from
below by $\epsilon{=}10^{-6}$:
\begin{equation*}
    R_k = \frac{\mathrm{MSE}_k}{\max(\hat{\sigma}_k^2,\, \epsilon)}.
\end{equation*}
In practice, $\hat{\sigma}_k^2$ never falls below $10^{-4}$ on
either dataset (B1: $2.2{\times}10^{-3}$ on SIDD), so this
safeguard is never triggered but ensures robustness in
out-of-distribution settings.

\noindent\textbf{Softmax Group DRO.} After noise normalization, bands still improve at different rates
throughout training: bright bands converge quickly while dark bands
lag persistently, and the gap between them changes at every step.
A static weight preset before training cannot respond to this
ongoing disparity; only a mechanism that recomputes weights at
every training step can continuously redirect gradient flow to
whichever band is currently most in need.

We approximate the non-differentiable minimax $\min_\theta\max_k R_k$
with differentiable softmax reweighting over valid bands
$\mathcal{V}{=}\{k:\sum_i\phi_{ik}>0\}$:
\begin{equation}
    \mathcal{L}_\text{BBRD} = \sum_{k\in\mathcal{V}} w_k \cdot R_k,
    \quad
    w_k = \frac{\exp(\eta R_k)}{\sum_{j\in\mathcal{V}}\exp(\eta R_j)},
    \label{eq:bbrd_loss}
\end{equation}
where $\eta{=}5$ uniformly across all backbones and datasets.

\noindent\textbf{Temperature $\eta$:}
As $\eta \to \infty$, the loss approaches the hard minimax
$\max_k R_k$; at $\eta{=}0$, it reduces to a uniform average of
$R_k$ (normalized MSE). The value $\eta{=}5$ achieves the balance
between worst-band focus and overall performance, and remains stable
across $\eta \in [0.5, 10]$ (\cref{sec:ablation}). Crucially, $w_k$
are recomputed at every training step: as a lagging band improves,
its weight decreases and gradient flow shifts to the next worst band
(\cref{fig:dynamics_weights,fig:dynamics_rk}).
\subsection{Per-Band Evaluation Protocol}
\label{sec:protocol}

\textbf{PSNR-D} and \textbf{PSNR-B} denote per-band PSNR on the
darkest and brightest regions, respectively.
For SIDD, we use $[0, 0.2)$ / $[0.8, 1.0]$; for PolyU, whose
brightness is heavily skewed toward bright values
(bright/dark ratio ${\approx}18.5{\times}$), we use
$[0, 0.45)$ / $[0.7, 1.0]$ to ensure sufficient dark-band
pixel counts.
\ours{} consistently outperforms MSE across all alternative
dark-band thresholds: on SIDD, gains range from
$+0.45$\,dB at $[0, 0.3)$ to $+0.58$\,dB at $[0, 0.15)$,
with the largest gain at the most extreme dark region,
confirming that \ours{} is most effective precisely where
brightness bias is most severe (full threshold sweep in supplementary).
\section{Experiments}
\label{sec:exp}

\noindent\textbf{Datasets.}
We evaluate on two complementary real-world benchmarks.
SIDD~\cite{sidd} consists of smartphone images skewed toward
dark values (\cref{fig:tradeoff}a), making it a natural testbed
for dark-band failures.
PolyU~\cite{polyu} provides DSLR images with an extreme
bright-to-dark energy ratio of $18.5{\times}$, a complementary
regime where dark pixels are sparse and easily overlooked under
aggregate evaluation.

\noindent\textbf{Architectures.}
To verify that brightness bias is a loss-level phenomenon rather
than a model-specific artifact, we train \numarch{} architectures
spanning CNNs (NAFNet~\cite{nafnet}, DRUNet~\cite{drunet},
KBNet~\cite{kbnet}, SCUNet~\cite{scunet}), Transformers
(SwinIR~\cite{swinir}, Restormer~\cite{restormer}), and State
Space Models (MambaIR~\cite{mambair}, MambaIRv2~\cite{mambairv2}).

\noindent\textbf{Training.}
All models are trained with identical settings (Adam, 100 epochs,
fp32; full hyperparameters in supplementary).
\ours{} requires \emph{no} per-backbone tuning, as $K$, $\eta$,
and $\sigma_g$ are fixed across all \numarch{} architectures
and both datasets. It introduces only $+1.7\%$ training overhead,
broken down as follows: GMM fitting and $\hat{\sigma}_k^2$ estimation
are performed \emph{once} before training (${\sim}2$\,min on SIDD);
per-step, the soft weights $\phi_{ik}$ are recomputed from the
clean target's brightness map (element-wise Gaussian, negligible),
and the softmax DRO adds $K$ scalar exponentials and one
normalization.
\ours{} incurs \textbf{zero inference cost}: band assignments and
DRO weights are discarded after training, leaving the deployed
model bitwise identical to one trained with vanilla MSE.
\ours{} operates in sRGB to match both datasets and all
evaluated architectures, with $\hat{\sigma}_k^2$ estimated
empirically from training pairs $(x,y)$, automatically
absorbing ISP nonlinearities without any parametric noise
assumption. $\hat{\sigma}_k^2$ is stable across subsamples
(CV ${<}15\%$), ISO subgroups (${<}8\%$), and cross-dataset
transfer (SIDD $\hat{\sigma}_k^2$ on PolyU yields PSNR-D
within 0.1\,dB).
A $2{\times}$ uniform scaling of $\hat{\sigma}_k^2$ changes PSNR-D by
${<}0.3$\,dB, consistent with the observation that DRO weights depend
on relative band ranking rather than absolute $\hat{\sigma}_k^2$
values (supplementary).

\noindent\textbf{Baselines.}
We compare against \numbaseline{} loss functions across four paradigms:
pixel-wise regression ($\ell_2$, $\ell_1$, Charbonnier~\cite{charbonnier1994},
Huber~\cite{huber1964}), structural similarity
(SSIM~\cite{ssim2004}, MS-SSIM~\cite{msssim}), frequency-domain
objectives (FFT~\cite{fftloss}, $\ell_1$+FFT~\cite{ntire2024lowlight},
Focal Freq.~\cite{focalfreq2020}, Guided Freq.~\cite{gfl2023}),
and sample reweighting (Focal~\cite{focalloss}, OHEM~\cite{ohem},
Heteroscedastic NLL~\cite{kendall2017}).

\subsection{Main Results}
\label{sec:main_results}
\cref{tab:main_loss} reveals a consistent failure: every baseline either preserves or worsens the dark/bright disparity, as raw residuals inflated by signal-dependent noise cause all pixel-wise objectives to systematically over-serve bright bands.
Notably, the closest prior work---Heteroscedastic NLL~\cite{kendall2017}---\emph{worsens} PSNR-D below MSE (37.29 vs.\ 37.51\,dB), as its learned per-pixel $\sigma(x)$ assigns low uncertainty to dark bands whose noise variance is genuinely low, reinforcing the bias. \ours{} addresses this by normalizing at the \emph{group level}, a distinction per-pixel methods cannot capture.

\ours{} breaks this trade-off: in our experiments it is the
\textbf{only method among all \numbaseline{} alternatives that
simultaneously improves PSNR-D and PSNR-B on both datasets}:
\textbf{+0.65\,dB} aggregate PSNR on SIDD
and \textbf{+0.56\,dB} on PolyU, with per-band gains
reaching up to \textbf{+0.89\,dB}, pushing the
PSNR-D vs.\ PSNR-B Pareto frontier strictly outward
(\cref{fig:pareto_sidd}).

\begin{figure*}[!t]
  \centering
  \includegraphics[width=0.9\linewidth]{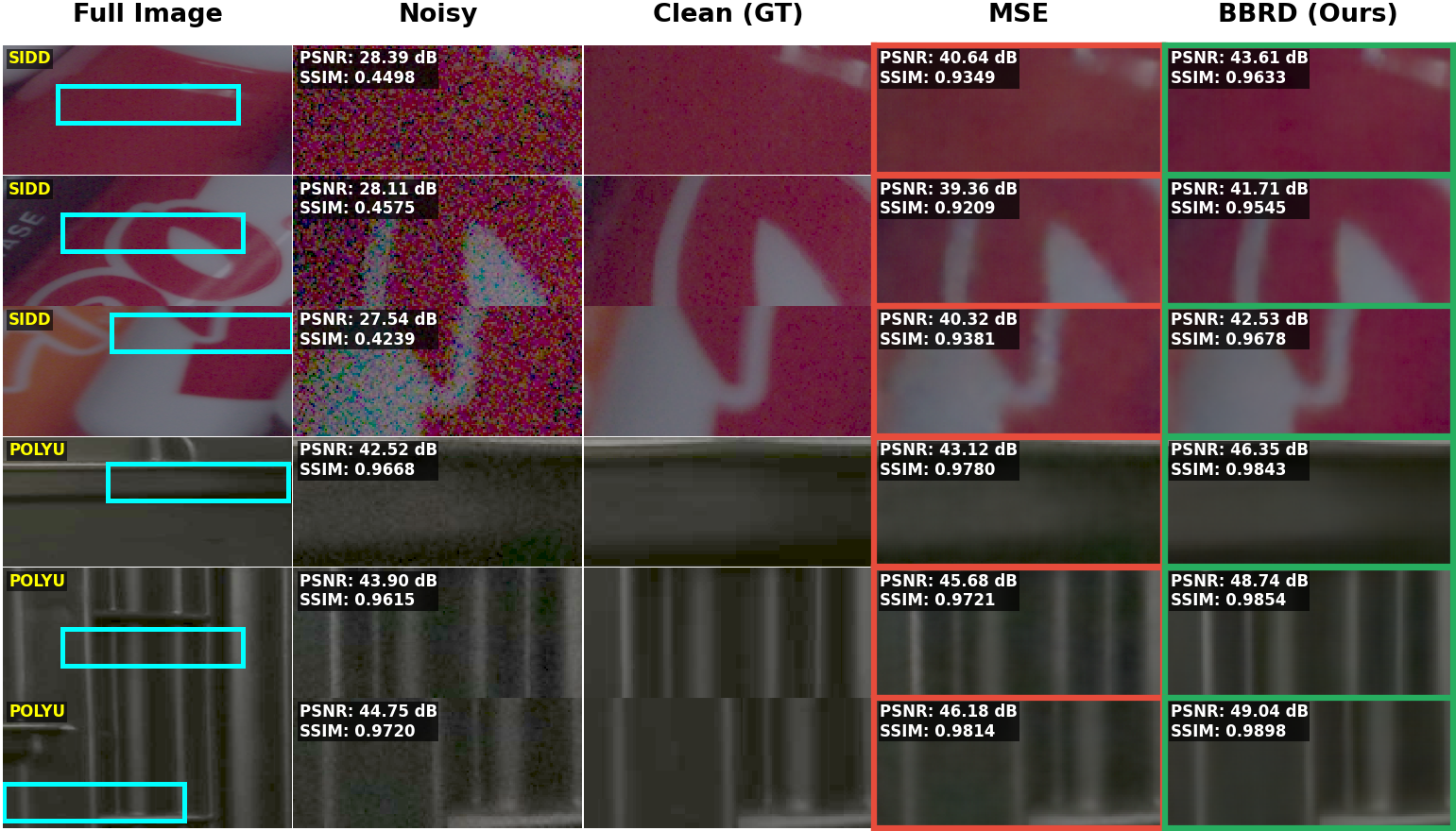}
  \caption{%
    \textbf{Qualitative comparison (SIDD and PolyU).}
    \ours{} corrects dark-region residuals left behind by MSE
    (green $=$ improvement; red $=$ regression, confined to
    bright areas where MSE already performs well).}
  \label{fig:qualitative}
\end{figure*}

\cref{tab:arch} confirms this is a loss-level bottleneck: \ours{} improves every metric on all \numarch{} backbones---CNN, Transformer, and SSM---with no per-backbone tuning.
Qualitative comparisons (\cref{fig:qualitative}) show \ours{} corrects dark-region residuals that MSE leaves behind.
\begin{table*}[!t]
  \centering
  \small
  \renewcommand{\arraystretch}{1}
  \caption{%
    \textbf{Loss comparison (NAFNet).}
    PSNR-D\,/\,PSNR-B: SIDD $[0,0.2)$\,/\,$[0.8,1.0]$;
    PolyU $[0,0.45)$\,/\,$[0.7,1.0]$.
    Every baseline either preserves or worsens the dark/bright
    disparity; \ours{} is the only method that improves both bands
    simultaneously.
    Best \textbf{bold}, second-best \underline{underlined}.}
  \label{tab:main_loss}
  \setlength{\tabcolsep}{5pt}
  \resizebox{\linewidth}{!}{%
  \begin{tabular}{@{}l cccc cccc@{}}
  \toprule
  & \multicolumn{4}{c}{\textbf{SIDD} (smartphone)}
  & \multicolumn{4}{c}{\textbf{PolyU} (DSLR)} \\
  \cmidrule(lr){2-5}\cmidrule(lr){6-9}
  Method
    & PSNR-D$\uparrow$ & PSNR-B$\uparrow$ & PSNR$\uparrow$ & SSIM$\uparrow$
    & PSNR-D$\uparrow$ & PSNR-B$\uparrow$ & PSNR$\uparrow$ & SSIM$\uparrow$ \\
  \midrule
  \multicolumn{9}{@{}l}{\textit{Pixel-wise regression}} \\
  MSE ($\ell_2$)
    & 37.51 & 38.37 & 39.47 & .9389
    & 35.80 & 37.82 & 38.50 & .9748 \\
  $\ell_1$
    & 37.60 & 37.67 & 39.65 & .9390
    & 35.94 & 37.87 & 38.68 & .9783 \\
  Charbonnier~\cite{charbonnier1994}
    & 37.48 & \underline{38.66} & 39.65 & .9398
    & 35.99 & 38.18 & 38.79 & .9794 \\
  Huber~\cite{huber1964}
    & 37.44 & 38.34 & 39.51 & .9400
    & 35.90 & 37.65 & 38.35 & .9754 \\
  \midrule
  \multicolumn{9}{@{}l}{\textit{Structural}} \\
  SSIM~\cite{ssim2004}
    & 36.81 & 36.49 & 38.78 & .9337
    & 36.07 & 38.05 & 38.85 & .9793 \\
  MS-SSIM~\cite{msssim}
    & 37.50 & 37.64 & 39.36 & .9396
    & 36.06 & 37.90 & 38.72 & .9779 \\
  \midrule
  \multicolumn{9}{@{}l}{\textit{Frequency-domain}} \\
  FFT~\cite{fftloss}
    & 37.64 & 37.96 & 39.59 & .9406
    & 35.84 & 37.56 & 38.44 & .9764 \\
  $\ell_1$+FFT~\cite{ntire2024lowlight}
    & \underline{37.70} & 38.28 & \underline{39.80} & \underline{.9417}
    & 36.11 & \underline{38.20} & 38.80 & .9786 \\
  Focal Freq.~\cite{focalfreq2020}
    & 36.10 & 37.07 & 37.67 & .9058
    & 35.20 & 37.15 & 37.54 & .9622 \\
  Guided Freq.~\cite{gfl2023}
    & 37.57 & 38.60 & 39.50 & .9389
    & \underline{36.15} & 38.14 & \underline{38.92} & \underline{.9802} \\
  \midrule
  \multicolumn{9}{@{}l}{\textit{Reweighting \& mining}} \\
  Focal~\cite{focalloss}
    & 37.19 & 38.55 & 39.16 & .9339
    & 35.54 & 37.48 & 38.07 & .9711 \\
  OHEM~\cite{ohem}
    & 37.47 & 38.64 & 39.44 & .9385
    & 36.08 & 37.92 & 38.62 & .9779 \\
  Heteroscedastic NLL~\cite{kendall2017}
    & 37.29 & 38.09 & 39.37 & .9386
    & 36.09 & 38.12 & 38.75 & .9788 \\
  \midrule
  \multicolumn{9}{@{}l}{\textit{Ours}} \\
  \ours{}
    & \textbf{37.96} & \textbf{38.69} & \textbf{40.12} & \textbf{.9442}
    & \textbf{36.27} & \textbf{38.34} & \textbf{39.06} & \textbf{.9804} \\
  \bottomrule
  \end{tabular}}
\end{table*}
\begin{table*}[!t]
\centering
\small
\renewcommand{\arraystretch}{1}
\setlength{\tabcolsep}{5pt}

\caption{%
  \textbf{Architecture generalization} (MSE vs.\ \ours{}).
  \ours{} improves all metrics on all \numarch{} backbones
  with no per-backbone tuning.
  PSNR-D\,/\,PSNR-B: SIDD $[0,0.2)$\,/\,$[0.8,1.0]$;
  PolyU $[0,0.45)$\,/\,$[0.7,1.0]$.}
\label{tab:arch}
\resizebox{\linewidth}{!}{
\begin{tabular}{@{}ll cccc cccc@{}}
\toprule
& & \multicolumn{4}{c}{\textbf{SIDD}}
  & \multicolumn{4}{c}{\textbf{PolyU}} \\
\cmidrule(lr){3-6}\cmidrule(lr){7-10}
Backbone & Loss
  & PSNR-D$\uparrow$ & PSNR-B$\uparrow$ & PSNR$\uparrow$ & SSIM$\uparrow$
  & PSNR-D$\uparrow$ & PSNR-B$\uparrow$ & PSNR$\uparrow$ & SSIM$\uparrow$ \\
\midrule
\multirow{2}{*}{NAFNet}
  & MSE     & 37.51 & 38.37 & 39.47 & .9389 & 35.80 & 37.82 & 38.50 & .9748 \\
  & \ours{} & \textbf{37.96} & \textbf{38.69} & \textbf{40.12} & \textbf{.9442}
            & \textbf{36.27} & \textbf{38.34} & \textbf{39.06} & \textbf{.9804} \\
\midrule
\multirow{2}{*}{DRUNet}
  & MSE     & 38.02 & 38.63 & 40.04 & .9445 & 35.92 & 37.98 & 38.50 & .9755 \\
  & \ours{} & \textbf{38.35} & \textbf{39.31} & \textbf{40.46} & \textbf{.9473}
            & \textbf{36.40} & \textbf{38.12} & \textbf{39.07} & \textbf{.9800} \\
\midrule
\multirow{2}{*}{KBNet}
  & MSE     & 37.62 & 38.48 & 39.82 & .9412 & 36.28 & 38.34 & 38.89 & .9749 \\
  & \ours{} & \textbf{38.43} & \textbf{38.76} & \textbf{40.41} & \textbf{.9462}
            & \textbf{36.47} & \textbf{38.77} & \textbf{39.00} & \textbf{.9793} \\
\midrule
\multirow{2}{*}{SwinIR}
  & MSE     & 37.29 & 39.27 & 39.48 & .9422 & 36.28 & 38.03 & 39.01 & .9792 \\
  & \ours{} & \textbf{37.83} & \textbf{39.77} & \textbf{40.23} & \textbf{.9450}
            & \textbf{36.33} & \textbf{38.29} & \textbf{39.23} & \textbf{.9798} \\
\midrule
\multirow{2}{*}{Restormer}
  & MSE     & 37.88 & 38.66 & 40.19 & .9437 & 36.33 & 38.54 & 39.01 & .9782 \\
  & \ours{} & \textbf{38.56} & \textbf{39.00} & \textbf{40.72} & \textbf{.9485}
            & \textbf{36.64} & \textbf{38.67} & \textbf{39.10} & \textbf{.9800} \\
\midrule
\multirow{2}{*}{SCUNet}
  & MSE     & 38.26 & 38.82 & 40.34 & .9461 & 35.96 & 38.08 & 38.68 & .9772 \\
  & \ours{} & \textbf{38.42} & \textbf{39.37} & \textbf{40.54} & \textbf{.9467}
            & \textbf{36.24} & \textbf{38.24} & \textbf{39.13} & \textbf{.9796} \\
\midrule
\multirow{2}{*}{MambaIR}
  & MSE     & 37.96 & 38.93 & 40.09 & .9432 & 36.16 & 38.09 & 38.77 & .9773 \\
  & \ours{} & \textbf{38.22} & \textbf{39.82} & \textbf{40.26} & \textbf{.9467}
            & \textbf{36.32} & \textbf{38.30} & \textbf{39.10} & \textbf{.9791} \\
\midrule
\multirow{2}{*}{MambaIRv2}
  & MSE     & 38.14 & 39.35 & 40.04 & .9449 & 36.02 & 38.11 & 38.75 & .9758 \\
  & \ours{} & \textbf{38.45} & \textbf{39.72} & \textbf{40.52} & \textbf{.9476}
            & \textbf{36.32} & \textbf{38.29} & \textbf{39.22} & \textbf{.9791} \\
\bottomrule
\end{tabular}}
\end{table*}

\section{Ablation Studies}
\label{sec:ablation}

\subsection{Noise Type Robustness}
\label{sec:noise_robustness}
We evaluate the SIDD-trained NAFNet on
\textbf{LumaSet-700}, a synthetic dataset of 700 patches
stratified uniformly across five brightness quintiles
($[0,0.2)$, $[0.2,0.4)$, \ldots, $[0.8,1.0]$), each containing
140 patches cropped from publicly available natural images and
rendered under three noise types (signal-dependent SIDD-style,
homoscedastic Gaussian, and random-variance) at 11 severity
levels (L0 mild $\to$ L10 max).
This is the only benchmark with uniform $[0,1]$ brightness
coverage and controllable noise. LumaSet-700 will be
released upon publication.
\ours{} outperforms MSE across all noise types and levels
(\cref{tab:lumaset}): the largest gain appears under
signal-dependent noise (+0.51\,dB), where variance normalization
is most principled, while gains persist even under homoscedastic
noise---confirming that \ours{} addresses both compounding sources
of brightness bias, not just noise heteroscedasticity.
Full per-level curves are in the
supplementary.

\begin{table}[tb]
\centering
\small
\caption{%
  \textbf{LumaSet-700 results} (SIDD-trained NAFNet, averaged over all 11 levels).
  \ours{} improves PSNR-D, PSNR-B, PSNR, and SSIM across all three noise types.
  PSNR-D\,/\,PSNR-B: $[0,0.2)$\,/\,$[0.8,1.0]$.
  Per-level curves are in the supplementary.}
\label{tab:lumaset}
\setlength{\tabcolsep}{3pt}
\resizebox{\linewidth}{!}{%
\begin{tabular}{@{}ll cccc cccc cccc@{}}
\toprule
& & \multicolumn{4}{c}{\textbf{Sig.-dep.}}
  & \multicolumn{4}{c}{\textbf{Random}}
  & \multicolumn{4}{c}{\textbf{Gaussian}} \\
\cmidrule(lr){3-6}\cmidrule(lr){7-10}\cmidrule(lr){11-14}
& Loss
  & PSNR-D$\uparrow$ & PSNR-B$\uparrow$ & PSNR$\uparrow$ & SSIM$\uparrow$
  & PSNR-D$\uparrow$ & PSNR-B$\uparrow$ & PSNR$\uparrow$ & SSIM$\uparrow$
  & PSNR-D$\uparrow$ & PSNR-B$\uparrow$ & PSNR$\uparrow$ & SSIM$\uparrow$ \\
\midrule
\multirow{2}{*}{NAFNet}
  & MSE     & 26.50 & 19.00 & 26.43 & .6536
            & 26.52 & 21.37 & 27.29 & .6887
            & 26.68 & 21.50 & 27.44 & .6931 \\
  & \ours{} & \textbf{26.80} & \textbf{19.37} & \textbf{26.94} & \textbf{.6870}
            & \textbf{26.81} & \textbf{21.76} & \textbf{27.71} & \textbf{.7123}
            & \textbf{26.98} & \textbf{21.89} & \textbf{27.87} & \textbf{.7172} \\
\bottomrule
\end{tabular}}
\end{table}

\subsection{Training Dynamics and DRO Stability}
\label{sec:training_dynamics}

\begin{figure*}[ht]
  \centering
  \includegraphics[width=0.99\linewidth]{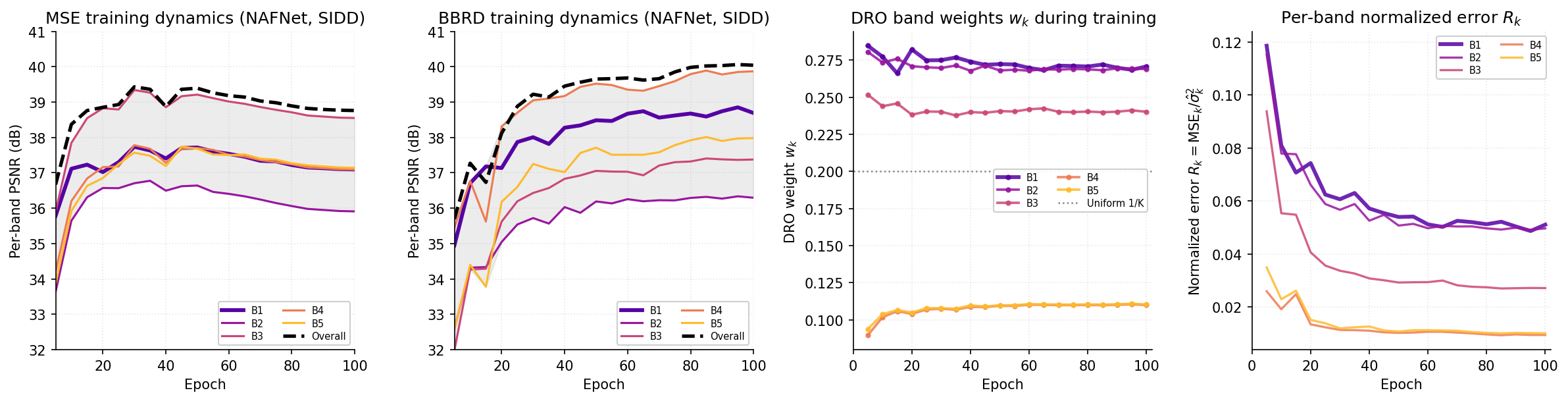}
  \caption{%
    \textbf{Training dynamics and DRO stability.}
    \textbf{(a)}~MSE locks in the dark/bright gap early.
    \textbf{(b)}~\ours{} closes this gap across all bands.
    \textbf{(c)}~DRO weights stabilize with B1 (dark) highest.
    \textbf{(d)}~$R_k$ decreases monotonically on a noise-corrected scale.}
\label{fig:theory_combined}
\label{fig:dynamics_mse}
\label{fig:dynamics_weights}
\label{fig:dynamics_rk}
\end{figure*}

\cref{fig:theory_combined} traces the full training trajectory.
Under MSE~(a), the dark/bright disparity locks in within the first
few epochs and never recovers.
\ours{}~(b) closes this gap from epoch~1, surpassing MSE's
worst-band ceiling on \emph{every} band by epoch~100
(gains: \textbf{+0.55\,dB} darkest $\to$ \textbf{+1.02\,dB} mid-tones).
Band weights $w_k$~(c) converge and stabilize with no oscillation;
normalized error $R_k$~(d) decreases monotonically, confirming that
DRO continuously tracks the truly lagging band on a noise-corrected
scale throughout training.

\subsection{Component Ablation}
\label{sec:component_ablation}
\cref{tab:ablation} traces a bottom-up path from MSE to full \ours{}.
Crucially, normalization must precede DRO: without it, DRO
\emph{worsens} PSNR-D below MSE (36.67\,dB${\downarrow}$) by
misidentifying high-noise bright bands as worst-performing.
Similarly, $\eta{=}0$ (normalization only) fails (36.82\,dB${\downarrow}$),
confirming both components are jointly necessary.
Band assignments from noisy $x$ rather than clean $y$ drop PSNR-D
by 0.53\,dB, and uniform quantile bands fall 0.05\,dB short,
validating our GMM-based clean-target design.
$K{=}5$ (BIC-selected) outperforms $K{=}4$ ($-$0.24\,dB) and
$K{=}8$ ($-$0.30\,dB); full sensitivity analysis is in the supplementary.
\begin{table*}[tb]
\centering
\small
\caption{%
  \textbf{Component ablation} (NAFNet, SIDD).
  DRO without normalization worsens PSNR-D below MSE; normalization
  must precede DRO. $\eta{=}0$ (normalization only) also fails,
  confirming both components are jointly necessary.
  Default in \colorbox{gray!15}{gray};
  ${\downarrow}$ denotes below MSE\@.
  PSNR-D\,/\,PSNR-B: $[0,0.2)$\,/\,$[0.8,1.0]$.}
\label{tab:ablation}
\setlength{\tabcolsep}{4pt}
\resizebox{\linewidth}{!}{%
\begin{tabular}{@{}llcccc@{}}
\toprule
Component & Setting
  & PSNR$\uparrow$ & SSIM$\uparrow$
  & PSNR-D$\uparrow$ & PSNR-B$\uparrow$ \\
\midrule
\multirow{4}{*}{Build-up}
  & MSE baseline
    & 39.47 & .9389 & 37.51 & 38.37 \\
  & DRO only (no norm, $\eta{=}5$)
    & 39.43 & .9402 & 36.67${\downarrow}$ & 38.44 \\
  & $+$ GMM bands, no $\hat{\sigma}_k^2$
    & 39.81 & .9420 & 37.07${\downarrow}$ & 38.50 \\
  & \cellcolor{gray!15}\textbf{Full \ours{} (GMM+Emp+Gaussian)}
    & \cellcolor{gray!15}\textbf{40.12} & \cellcolor{gray!15}\textbf{.9442}
    & \cellcolor{gray!15}\textbf{37.96} & \cellcolor{gray!15}\textbf{38.69} \\
\midrule
\multirow{3}{*}{Simple baselines}
  & Per-pixel inv.\ variance weighting
    & 39.11 & .9382 & 37.35${\downarrow}$ & 38.21${\downarrow}$ \\
  & Uniform quantile bands $+$ norm $+$ DRO
    & 39.97 & .9426 & 37.91 & 38.70 \\
  & \cellcolor{gray!15}\textbf{GMM bands $+$ norm $+$ DRO (ours)}
    & \cellcolor{gray!15}\textbf{40.12} & \cellcolor{gray!15}\textbf{.9442}
    & \cellcolor{gray!15}\textbf{37.96} & \cellcolor{gray!15}\textbf{38.69} \\
\midrule
\multirow{2}{*}{Band source}
  & Band from noisy $x$
    & 39.03 & .9381 & 37.43${\downarrow}$ & 38.40${\downarrow}$ \\
  & \cellcolor{gray!15}\textbf{Band from clean $y$ (ours)}
    & \cellcolor{gray!15}\textbf{40.12} & \cellcolor{gray!15}\textbf{.9442}
    & \cellcolor{gray!15}\textbf{37.96} & \cellcolor{gray!15}\textbf{38.69} \\
\midrule
\multirow{3}{*}{DRO temperature $\eta$}
  & $\eta{=}0$ \textit{(norm only, no DRO)}
    & 39.53 & .9373 & 36.82${\downarrow}$ & 38.51 \\
  & $\eta{=}1$
    & 40.04 & .9437 & 36.80${\downarrow}$ & 38.67 \\
  & \cellcolor{gray!15}$\boldsymbol{\eta{=}5}$ \textbf{(ours)}
    & \cellcolor{gray!15}\textbf{40.12} & \cellcolor{gray!15}\textbf{.9442}
    & \cellcolor{gray!15}\textbf{37.96} & \cellcolor{gray!15}\textbf{38.69} \\
\bottomrule
\end{tabular}}
\end{table*}

\section{Conclusion}
\label{sec:conclusion}


We propose \ourslong{} (\ours{}), a plug-and-play module that tackles an overlooked problem in image denoising tasks: dark pixels are reconstructed substantially worse than the bright ones despite carrying less signal-dependent noise. Specifically, \ours{} contains three stages. First, GMM band partitioning with Gaussian soft assignments identifies brightness-specific failure modes in a data-adaptive manner. Second, empirical noise normalization places all bands on a noise-corrected scale, ensuring the truly underperforming band (not merely the noisiest) is correctly identified. Finally, softmax Group DRO dynamically redirects gradient flow to whichever band is currently lagging throughout training, with zero additional parameters or inference cost. Experiments across \numarch{} architectures and both datasets demonstrate that, in our experiments, \ours{} is the only objective among \numbaseline{} alternatives that improves every brightness band simultaneously, delivering up to \textbf{+0.45\,dB} on dark bands, \textbf{+0.32\,dB} on bright bands, and \textbf{+0.65\,dB} aggregate PSNR on SIDD (NAFNet).

\section*{Acknowledgements}
This work was supported by research grants from the RGC
(Project Nos.\ 16500825, R6005-24, and C6088-25Y), a research grant from the
Joint Research Scheme (JRS) under the National Natural Science Foundation of
China (NSFC) and the Research Grants Council (RGC) of Hong Kong
(Project No.\ N\_HKUST654/24), and a research grant from the Innovation and
Technology Commission (ITC) (Project No.\ GHP/124/22).

\bibliographystyle{splncs04}
\bibliography{main}
\end{document}